\documentclass{article}

\usepackage{times}
\usepackage{epsfig}
\usepackage{graphicx}
\usepackage{amsmath}
\usepackage{amssymb}
\usepackage{algorithm}
\usepackage{algorithmic}
\usepackage{setspace}
\usepackage{makecell}
\usepackage{booktabs}

\usepackage{spconf,amsmath,graphicx}

\title{Occluded pedestrian detection with visible IoU and box sign predictor}
\name{Ruiqi Lu, Huimin Ma$^*$\thanks{$^*$Corresponding author, Tel: +86-010-62781432, E-mail: mhmpub@mail.tsinghua.edu.cn. This work is supported by National Key Basic Research Program of China (No. 2016YFB0100900) and National Natural Science Foundation of China (No. 61773231).}}
\address{Department of Electronic Engineering, Tsinghua University, Beijing, 100084}
\begin{document}
%
\maketitle
\begin{abstract}
Training a robust classifier and an accurate box regressor are difficult for occluded pedestrian detection. 
Traditionally adopted Intersection over Union (IoU) measurement does not consider the occluded region of the object and leads to improper training samples. To address such issue, a modification called visible IoU is proposed in this paper to explicitly incorporate the visible ratio in selecting samples. 
Then a newly designed box sign predictor is placed in parallel with box regressor to separately predict the moving direction of training samples. It leads to higher localization accuracy by introducing sign prediction loss during training and sign refining in testing. Following these novelties, we obtain state-of-the-art performance on CityPersons benchmark for occluded pedestrian detection. 
\end{abstract}

\section{Introduction}
\label{sec:intro}
\begin{figure}[htb]
	
	\begin{minipage}[b]{.48\linewidth}
		\centering
		\centerline{\includegraphics[width=4.0cm,height=4.0cm]{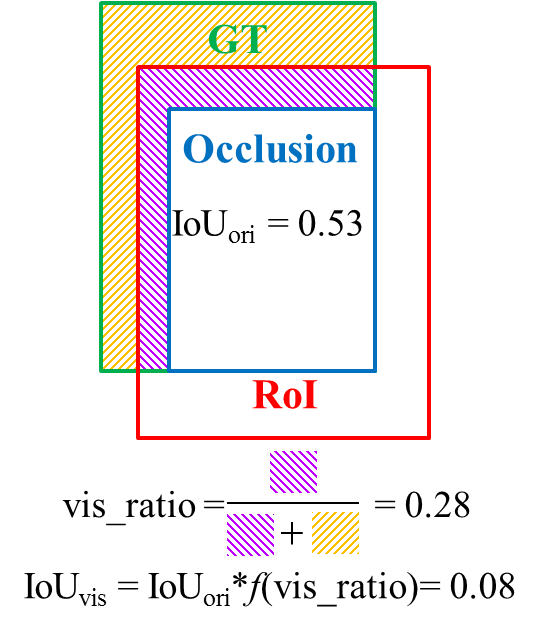}}
		\vspace{-1mm}
		\centerline{(a) Selecting positive samples.}\medskip
	\end{minipage}
	\hfill
	\begin{minipage}[b]{0.48\linewidth}
		\centering
		\centerline{\includegraphics[width=4.5cm, height=4.0cm]{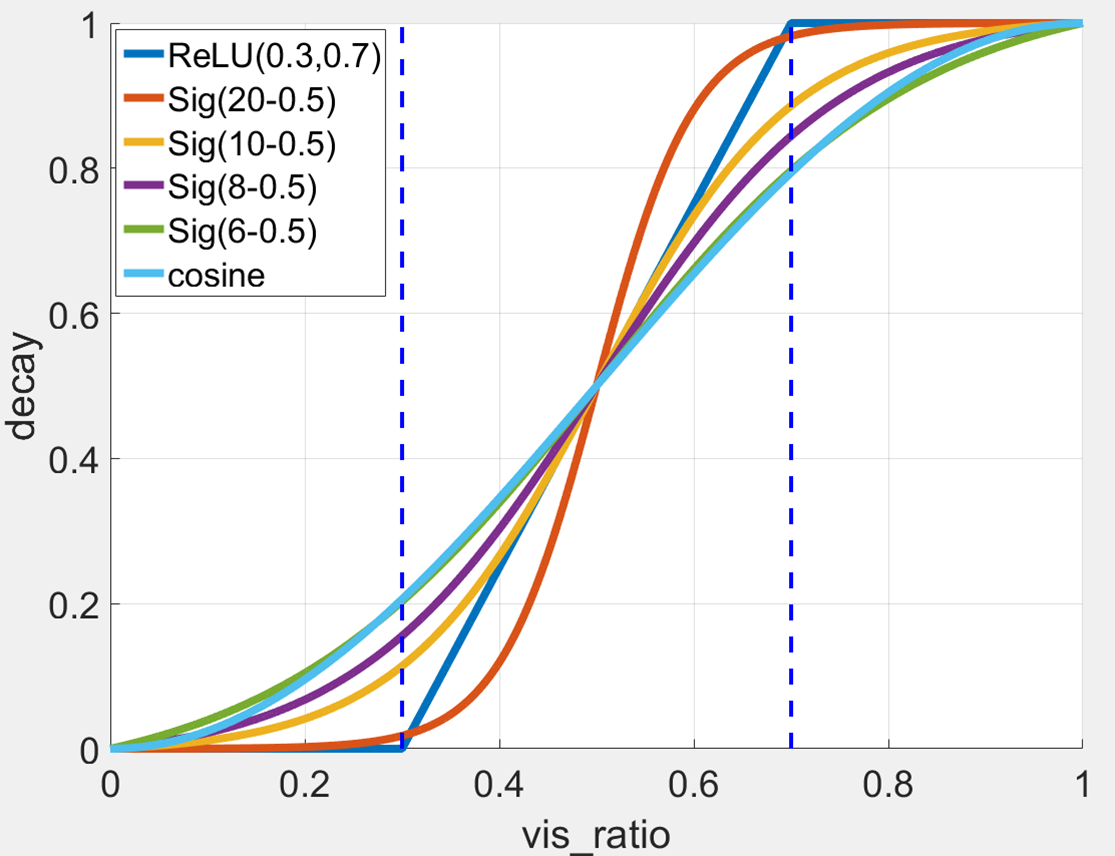}}
		\vspace{-1mm}
		\centerline{(b) Decay functions.}\medskip
	\end{minipage}
	\vspace{-4mm}
	\caption{Two ways of calculating IoU for selecting positive samples and visualization of decay functions. (a) When an object (green box) is occluded (blue box), a region of interest (red box) with IoU higher than a threshold (0.5) may contain large area of occlusion. With visible IoU, it can be easily discarded from positive training set. (b) Decay functions used to calculate visible IoU. ReLU($\rm x_1,x_2$): linearly increasing from ($\rm x_1$,0) to ($\rm x_2$,1). Cosine: -0.5cos($\pi*\rm x$)+0.5.}
	\label{fig:decay function}
	\vspace{-2mm}
	
\end{figure}

Despite being intensively studied in recent years, pedestrian detection under road scene still remains a challenging task because of occlusion. There are two aspects of difficulties in training and detecting occluded pedestrians. Firstly, the bounding box of an occluded object contains regions of others, making it hard to distinguish positive samples from negative ones. Shown in Fig.\ref{fig:decay function} (a), though the IoU between the ground truth box and the region of interest (RoI) is higher than the threshold (0.5), the RoI should not be regarded as a positive sample because most area in the RoI is occlusion. Therefore the visible ratio of the RoI should be taken into consideration when deciding its label. Secondly, as boxes of different pedestrians frequently overlap with each other, an inaccurately localized box will have subsequent impacts on other detections through Non-Maximum Suppression (NMS). Thus, producing precise boxes is crucial for crowd detection. Traditionally adopted box regressor in CNN based detectors \cite{fast,faster,mask}  needs carefully designed loss to guarantee stable performance. 
However, because the box regressor simultaneously produces the shifting direction and steps, it's possible to detach direction prediction from box regressor and achieve better convergence and localization accuracy.

The sampling of positive and negative training samples can cause great influences on the performance of classifier. SNIP \cite{snip} proposes a scale normalization method for training with image pyramids. It only selects the proper samples which fall in the desired scale range under different pyramids for training. Cascade R-CNN \cite{cascade} adopts cascaded classifiers where training samples with increasingly higher overlap with ground truths are fed. Online hard example mining (OHEM) \cite{ohem} dynamically chooses the samples with the highest loss in a batch to achieve better convergence and performance. While these methods are for general objects, we propose a sampling measurement specially designed for training with occluded objects, which considers explicitly the visible ratio of a RoI in choosing positive samples and is named as visible IoU.   

Most prior works targeted on refining objects detection focus on extracting and utilizing more discriminative features. They rely on designing complex network architectures and fusion operations and the down stream box regressor is hardly given attention to. Zhai~et~al.~\cite{feature-selective} propose a feature selective method to selectively pool different aspect ratio and sub-region features dependent on the characteristics of objects. R-FCN \cite{R-FCN} uses position-sensitive RoI pooling to introduce translation-variance in object detection. Following position pooling, MaskLab \cite{masklab} further proposes direction pooling to acquire more sensitive features. For pedestrian detection, human part feature \cite{part-1,part-2,part-3,ORCNN} is introduced to handle occlusion.
In \cite{fusedDNN, illuminating, small}, segmentation feature is adopted to further boost detection performance.
RPN+BF \cite{RPN+BF} replaces the second stage classifier with boosted forests. While more discriminative features indeed help producing better detections, the network grows large and inefficient, and different models are not compatible with each other. 

Different from prior works which mainly focus on the early stages of feature extraction and adopt complex network structure, we propose box sign predictor which aims at separately predicting the moving direction of boxes. It helps to train a better box predictor and improves the localization accuracy. Besides, it is light weight and can be easily incorporated in any other frameworks that adopt box regressors.

The main contributions of this paper are summarized as:

(1) We propose visible IoU, which explicitly considers the visible area in selecting positive samples to produce high quality training set for the classifier.

(2) We design box sign predictor and box sign prediction loss at the final stage to predict the moving direction of an RoI and improve the localization accuracy. 

(3) By introducing the two methods, we are able to achieve state-of-the-art performance on CityPersons occluded pedestrian detection benchmark.

\vspace{-1mm}
\section{Training with visible region}
\label{sec:vis}
\vspace{-1mm}
\subsection{Visible IoU}
\vspace{-1mm}
Traditional IoU measurement which is used to distinguish positive samples from negative ones places same importance on the visible and occluded region. It leads to improper positive training samples that only contain a little amount of visible area. To alleviate the problem, one optional way is increasing the IoU threshold, for example, from 0.5 to 0.7. As the IoU increases, there is more chance for an RoI to contain larger visible area. But the number of positive samples will decline drastically and not be enough to train a robust classifier. 

Different from the above method, the visible ratio of an RoI is considered in our method to calculate a more representative value, which is called visible IoU. It is defined as:
\vspace{-2mm}
\begin{equation}
{\rm IoU_{vis}} = {\rm IoU_{ori}} *f({\rm vis\_ratio})
\end{equation}
where $\rm vis\_ratio = \frac{|RoI\;\cap\;V|}{|V|}$. $\rm V$ represents the visible region of the ground truth box associated with the RoI, $\cap$ is the intersection of regions and  $|\cdot|$ represents the area within the region. Therefore, $\rm vis\_ratio$ is the fraction of visible area that is within the RoI. $f({\rm x})$ represents the decay function determined by $\rm vis\_ratio$. In this way, the visible IoU is able to characterize the visible area within an RoI while calculating the overall overlap.
\vspace{-1mm}
\subsection{Decay Function}
\vspace{-1mm}
The decay function should be monotonically increasing with $\rm vis\_ratio$. Inspired by the activation function adopted by modern convolutional neuron networks \cite{resnet, VGG}, the sigmoid form of decay function is used in the experiments, which is defined as:
\vspace{-2mm}
\begin{equation}
f_{sigmoid}({\rm x}) = \frac{s({\rm x})-s(0)}{s(1)-s(0)}, s({\rm x}) = \frac{1}{1+{\rm exp}^{-\beta({\rm x}-\alpha)}}
\vspace{-1mm}
\end{equation}
where $\alpha$ and $\beta$ are the hyper parameters of sigmoid function and $f(\rm x)$ is denoted as Sigmoid($\beta$, $\alpha$). Shown in Fig.\ref{fig:decay function} (b), $\alpha$ is fixed at 0.5 in this paper for symmetry. As $\beta$ increases, the sigmoid function becomes steeper near 0.5. Therefore, at the same $\rm vis\_ratio$, the decay is larger for larger $\beta$ and ${\rm IoU_{vis}}$ is more close to ${\rm IoU_{ori}}$, leading to more positive samples. 

Other form of functions, like ReLU and cosine are also tested in the experiments for comparison. The visualization of other functions can also be seen in Fig.\ref{fig:decay function} (b).

\begin{figure}
	\begin{minipage}[b]{.48\linewidth}
	\centering
	\centerline{\includegraphics[width=4.0cm, height=3.5cm]{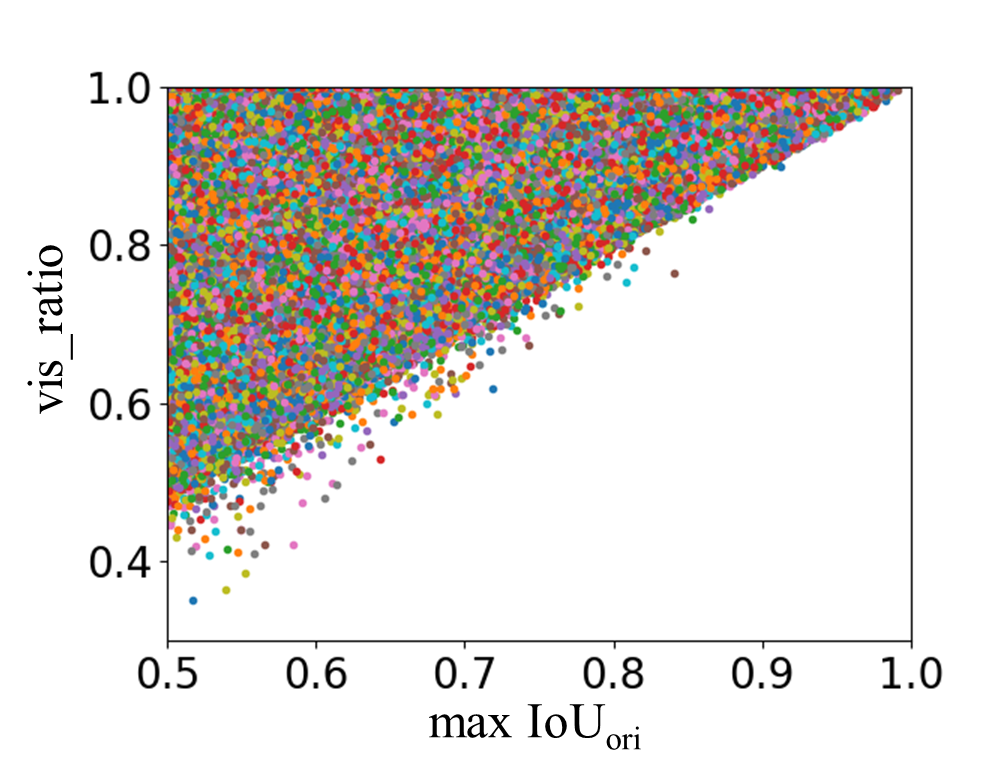}}
	\vspace{-2mm}
	\centerline{(a)\,No decay.}\medskip
\end{minipage}
\hfill
\begin{minipage}[b]{0.48\linewidth}
	\centering
	\centerline{\includegraphics[width=4.5cm,height=3.5cm]{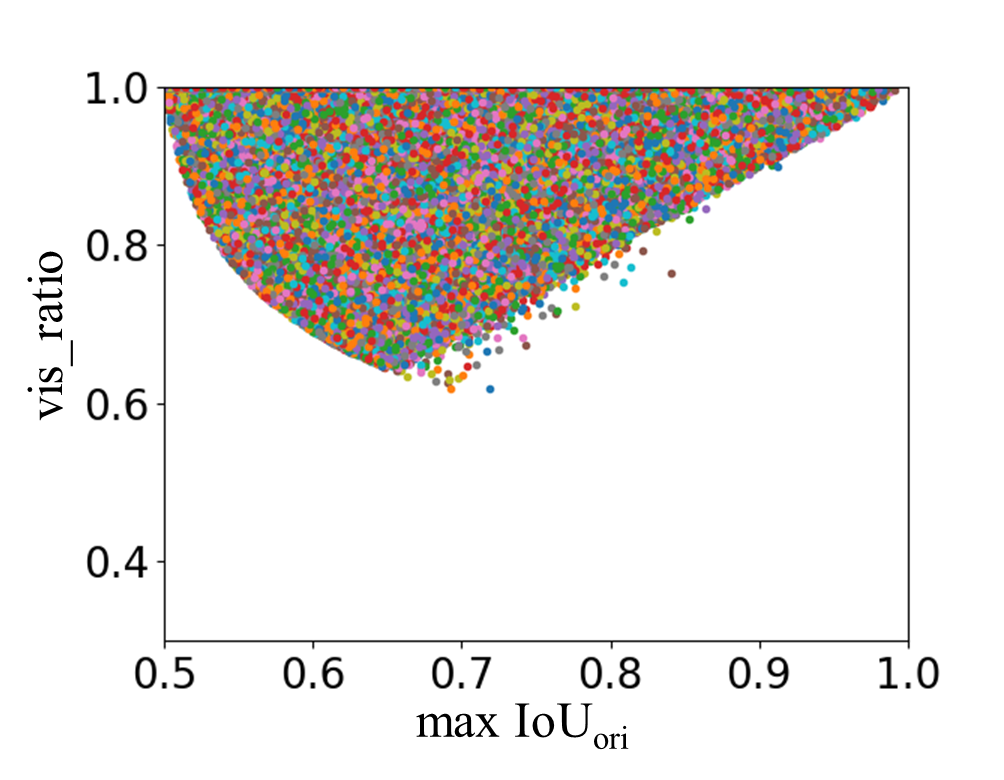}}
	\vspace{-2mm}
	\centerline{(b)\,Sigmoid(8,0.5).}\medskip
\end{minipage}
\vspace{-4mm}
	\caption{The distribution of positive samples in the $\rm vis\_ratio$-${\rm max\;IoU_{ori}}$ diagram, where ${\rm max\;IoU_{ori}}$ represents the maximum $\rm IoU_{ori}$ an RoI has with ground truth objects and is referred as $\rm IoU_{ori}$ for simplicity. Every dot represents a positive sample. Different colors represent different images. (a) The positive samples are selected using ${\rm IoU_{ori}}$ measurement. (b) Using ${\rm IoU_{vis}}$ measurement.}
	\label{fig:distribution}
	\vspace{-3mm}
\end{figure}

To better understand the influences of using visible IoU, the distribution of positive samples during training is plotted in Fig.\,\ref{fig:distribution}. We set the threshold for IoU as 0.5 and plot the positive samples in the top scored 100 RoIs of each training image. Comparing $\rm IoU_{ori}$ with $\rm IoU_{vis}$, most positive samples whose $\rm IoU_{ori}$ with a ground truth is higher than 0.7 are still positive after decay. The discarded samples concentrate at bottom left area whose $\rm vis\_ratio$ are low. Despite discarding many low quality samples, there are still enough ones comparing with setting the threshold as 0.7. 

\begin{figure}
	\begin{center}
		\includegraphics[width=1.0\linewidth,height=0.5\linewidth]{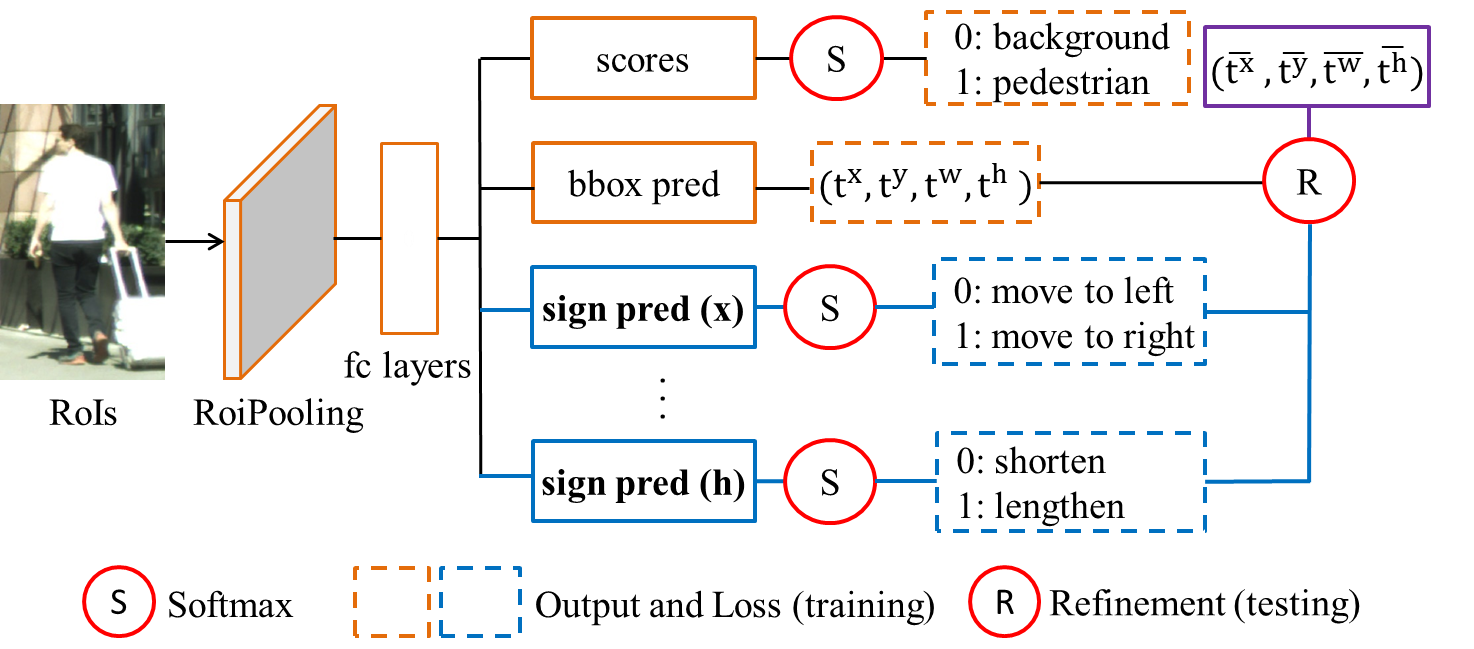}        
	\end{center}
	\vspace{-5mm}
	\caption{The structure of the framework. The box sign predictor is placed in parallel with classifier and regressor. The probability distribution of four box elements are predicted through fc layers and a softmax layer. Then the sign prediction results are further used to refine box predictions in test phase.}
	\label{fig:sign predictor}
	\vspace{-3mm}
\end{figure}

\vspace{-1mm}
\section{box sign prediction}
\label{sec:sign}
Localization accuracy is a critical problem under crowd occlusion. Normally the positive RoIs are roughly located near the ground truth objects and need refinement. Common models adopt bounding box regression to predict the proportional change of location and length with regard to original RoIs. However, it is a hard task to simultaneously predict the moving direction and steps, making the box regressor an unstable module to train. Therefore, we propose to detach the direction prediction from box regression.

Apart from common category classifier and bounding box regressor, we propose another submodule to further refine the localization accuracy, which is named as box sign predictor. Shown in Fig.\,\ref{fig:sign predictor}, the feature of a region of interest (RoI) is first extracted by RoIPooling layer. Then it is forwarded through several fully connected layers (fc6 and fc7 for VGG-16) before finally being sent to the classifier or regressor. The classifier outputs the probability of the RoI being background and a pedestrian. The box regressor outputs normalized offsets $(t^x, t^y, t^w, t^h)$ of position X, Y, width and height. 

The box sign predictor takes the same fc features as input and outputs the probability distribution for the four elements of box prediction results. Specifically, for the X position, the sign predictor outputs a two dimensional vector $ s^x = (s^{x-}, s^{x+})$, where $ s^{x-}$ and $s^{x+}$ represent the probability of the box moving to the left and right, respectively. For width of the box, the sign predictor also outputs $s^w = (s^{w-}, s^{w+})$, where $s^{w-}$ and $s^{w+}$ are the probability of decreasing and increasing the width, respectively. The sign predictors for Y position and height are similarly defined. With the box sign predictors, the overall loss for the detector becomes:
\vspace{-1mm}
\begin{equation}
	L_{detector} = L_{cls} + L_{box} + L_{sign}
\vspace{-1mm}
\end{equation}
where $L_{cls}$ and $L_{box}$ are the classification and regression loss and follow traditional SoftmaxLoss and SmoothL1Loss. $L_{sign}$ is the sign prediction loss and is defined as:
\vspace{-1mm}
\begin{equation}
\begin{split}
L_{sign}=\frac{\gamma}{N_{reg}}\sum_{k\in\{x,y,w,h\}}\sum_{i}&-[1(t_i^{k*}\leq0)log(s^{k-}_i)\\&+1(t^{k*}_i>0)log(s^{k+}_i)]
\end{split}
\end{equation}
where $N_{reg}$ is the total number of positive samples, k represents the dimension of box prediction, i is the index of positive RoIs in a mini-batch, and $t^{k*}$ represents the bounding box regression target for corresponding dimension $k$. $\gamma$ is the weight of $L_{sign}$ and is chosen manually to balance the magnitude of different loss. Note that $L_{sign}$ is essentially SoftmaxLoss and can be easily implemented. 

In the test phase, the box prediction results can be further refined by the sign prediction results through the following equation:
\vspace{-1mm}
\begin{equation}
	\overline{t^k} = 1(t^k\leq0)t^k \cdot s^{k-} + 1(t^k>0)t^k\cdot s^{k+}
\vspace{-1mm}
\end{equation}

The refinement operation can further prevent the box from shifting too far or to the wrong direction and refine the localization accuracy.  

\begin{table*}
	\begin{center}
		\renewcommand\arraystretch{1.2}
		\begin{tabular}{@{}lcccccccc@{}}
			\Xhline{1.0pt}
			Method & backbone & scale& Reasonable & Heavy & Partial & Bare & Reasonable-test & Heavy-test \\
			\hline
			Adapted Faster-RCNN \cite{citypersons} & VGG-16 & $\times$ 1.3 & 12.8 & - & - & - & 12.97 & 50.47 \\
			Repulsion Loss \cite{repulsion} & ResNet-50 & $\times$ 1.3 & 11.6 & 55.3 & 14.8 & 7.0 & - & - \\
											&			& $\times$ 1.5 & 10.9 & 52.9 & 13.4 & 6.3 & 11.48 & 52.59\\   
			OR-CNN \cite{ORCNN} &VGG-16 & $\times$ 1.3& 11.0 & \bf{51.3} & 13.7 & \bf{5.9} & 11.32 & 51.43 \\
			\hline
			$\rm IoU_{vis}+Sign$(ours) & VGG-16 &$\times$ 1.3 & \bf{10.8} & 54.3 & \bf{10.6} & 7.0 & \bf{11.24} & \bf{45.02} \\
			\Xhline{1.0pt}
		\end{tabular}
	\end{center}
	\vspace{-4mm}
	\caption{Comparisons of the proposed model with other state-of-the-art methods. Scale means that the input images are up-sampled certain times indicated by the number while training and testing. Other than "reasonable" setup, the performance on Heavy, Partial, and Bare subsets are also provided. The last two columns are the performance on the test set. }
	\label{compare}
	\vspace{-2mm}
\end{table*}

\vspace{-1.5mm}
\section{experiments}
\vspace{-1mm}
\label{sec:expt}
\noindent \textbf{Dataset.} As the visible region is required in calculating visible IoU, there should be annotations for it in the dataset. Therefore, CityPersons \cite{citypersons} is chosen for evaluation. It comes from the semantic segmentation dataset CityScapes \cite {cityscape} and consists of 2975 training images and 500 validation images. Both bounding box annotations for the whole body and the visible parts are provided. The annotations for test set are not released and only used for online evaluation. We follow the evaluation metric MR$^{-2}$ (lower is better) calculated on the "reasonable" setup (height$>$50, occlusion$\leq0.35$) of the dataset.

\noindent \textbf{Implementation Details.} We implement the proposed visible IoU and box sign predictor on the basis of adapted Faster-RCNN framework \cite{faster, citypersons}. VGG-16 is used as the backbone network. On the CitypPersons benchmark, we train the network for 30k iterations with the base learning rate of 0.001 and decreased by a factor of 10 after 
the first 20k iterations. OHEM is used to accelerate convergence. The threshold of IoU or visible IoU is fixed at 0.5 throughout the experiments. The weight of sign prediction loss $\gamma$ is set as 0.1 to keep the balance with classification loss and box regression loss.  

\noindent \textbf{Ablation Studies.}
Table \ref{ablation study} (a) demonstrates the performance under different decay functions and hyper parameters. By introducing the visible IoU, the detection results improve gradually as $\beta$ decreases from 20 to 8 because more low quality positive samples are discarded. However, if we continue to decrease $\beta$, some of the high quality positive samples are also discarded and the performance declines. Other form of decay functions are also tested and show considerable improvement, which proves that our method does not depend on the explicit form of decay function. The best result decreases the MR$^{-2}$ by 0.57\% at Sigmoid(8,0.5). 

Table \ref{ablation study} (b) shows the performance when training with extra sign predictor. Visible IoU calculated by Sigmoid(8, 0.5) decay function is adopted in this experiment. The extra loss introduced by sign predictor decreases the MR$^{-2}$ by 0.60\%. By further conducting the refinement procedure introduced in equation (5), the performance reaches 12.96\%. Though the sign prediction loss indeed helps improving the performance, one can argue that it is because the loss involved with box prediction is increased and the sign predictor structure is not necessary. To prove that the form of sign predictor is essential for the improvement, we remove the sign predictor and increase the bounding box regression loss instead. Either sigma or the weight of SmoothL1Loss is increased to stress the importance of bounding box regression. However, it brings little improvement to the performance and therefore shows that to detach sign prediction from box regression is essential for the improvement.

\begin{figure}
	\begin{center}
		\includegraphics[width=1.0\linewidth,height=0.8\linewidth]{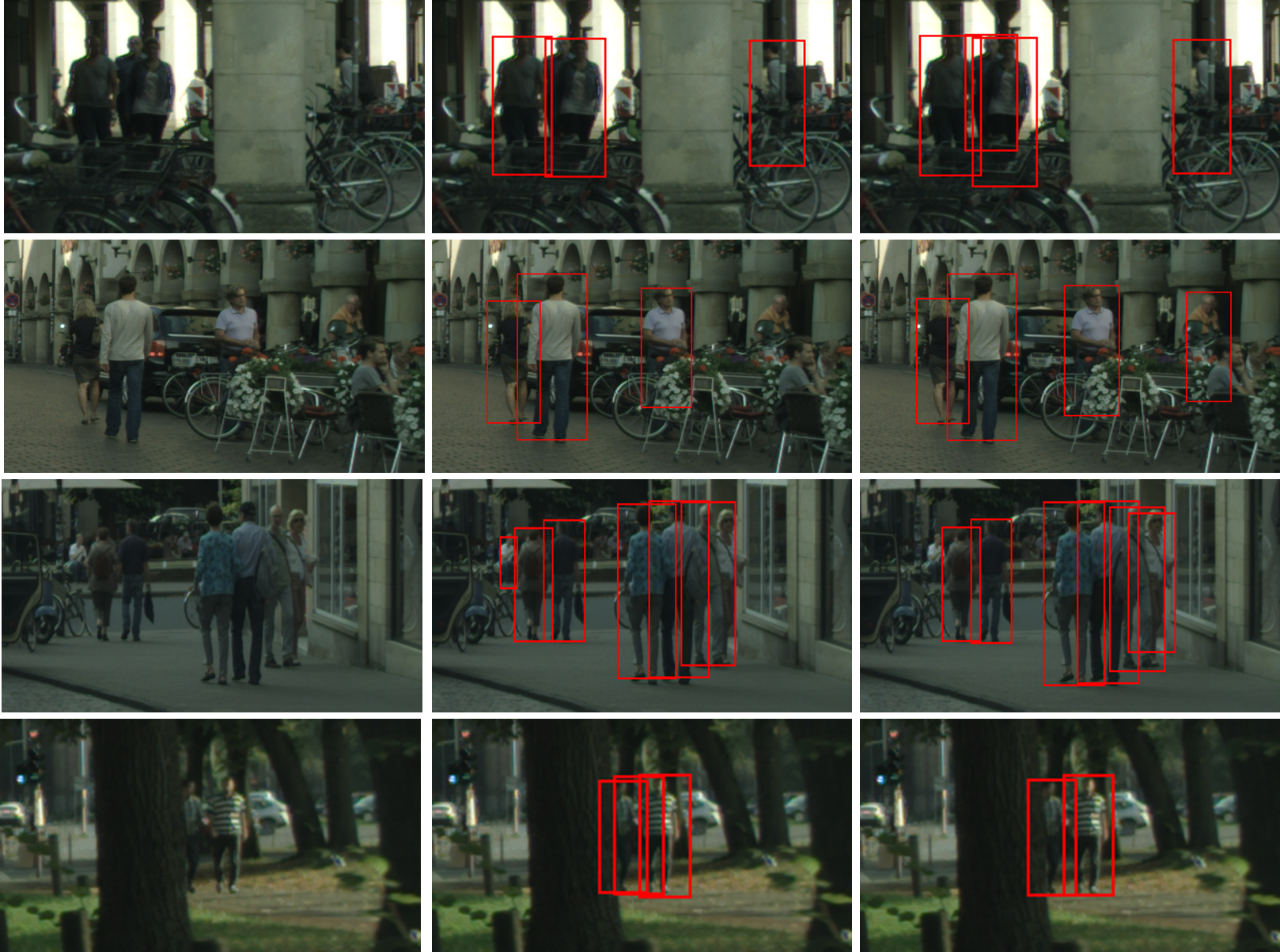}        
	\end{center}
	\vspace{-3mm}
	\centerline{\quad(a) input\quad\quad\quad\quad(b) baseline\quad\quad\quad(c) our model.}
	\vspace{-2.5mm}
	\caption{Visual comparisons between baseline and the complete model. Compared with baseline, our model outputs more accurate boxes and detects pedestrians with heavy occlusion.}
	\label{fig:visualizition}
	\vspace{-4mm}
\end{figure}

Some visual comparisons between the baseline and complete model are shown in Fig.\,\ref{fig:visualizition}. Our model produces more accurately localized boxes and detects pedestrians with heavy occlusion in a crowd. 

\vspace{-1mm}

\noindent \textbf{Comparisons with state-of-the-art methods.}
We report the comparisons between complete model and other methods on the validation and test set of CityPersons in Table.\,\ref{compare}. Following the split strategy in \cite{repulsion}, the reasonable subset is further divided into partial (10\% $<$ occlusion $\leq$ 35\%) and bare (occlusion $\leq$ 10\%) subsets. Heavy subset contains those annotations whose occlusion is above 35\% (which are not in the reasonable set).
Both OR-CNN \cite{ORCNN} and our model adopt visible region annotations, yet our network does not use human part features and is more lightweight. We achieve state-of-the art performance on the dataset. Specially, we surpass other methods on the partial and heavy-test subset with a large margin, which demonstrates the effectiveness of the proposed method under occlusion.
 
\begin{table}
	\begin{minipage}[b]{.50\linewidth}
		\centering
		\centerline{
			\renewcommand\arraystretch{1.1}
			\begin{tabular}{@{}lcc@{}}
				\Xhline{1.0pt}
				Decay Function & Reasonable  \\
				\hline
				None & 14.24 \\
				\hline
				Sigmoid(20)& 14.05 \\
				Sigmoid(10)& 13.77  \\
				Sigmoid(8)& \bf{13.67}\\ 
				Sigmoid(6)& 14.67\\
				\hline
				ReLU(0.3,0.7) & 13.89\\
				cosine & 14.07 \\
				\Xhline{1.0pt}
			\end{tabular}
		}
		\vspace{1mm}
		\centerline{(a) decay functions.}\medskip
	\end{minipage}
	\hfill
	\begin{minipage}[b]{.40\linewidth}
		\centering
		\centerline{
			\renewcommand\arraystretch{1.1}
			\begin{tabular}{@{}lcc@{}}
				\Xhline{1.0pt}
				Loss Change & Reasonable  \\
				\hline
				None & 13.67 \\
				\hline
				$\sigma=3$ & 13.62\\
				$\sigma=5$ & 14.80\\
				$\eta=2$ & 14.26\\
				$\eta=3$ & 14.46\\
				\hline
				+ sign loss& 13.07\\ 
				+ refining& \bf{12.96} \\
				\Xhline{1.0pt}
			\end{tabular}
		}
		\vspace{1mm}
		\centerline{(b) box sign prediction}\medskip
	\end{minipage}
	\vspace{-5mm}
	\caption{Ablation studies of the proposed model. (a)\,The $\alpha$ is fixed at 0.5 for sigmoid function. The curve for above functions can be seen in Fig.\,\ref{fig:decay function}. (b)\,The $\sigma$ and $\eta$ represent the sigma parameter and loss weight of SmoothL1Loss, respectively. The default value for both $\sigma$ and $\eta$ are 1.0 if not marked. }
	\label{ablation study}
	\vspace{-3mm}	
\end{table}
 
\vspace{-2mm}
\section{Conclusions}
\vspace{-2mm}
The occluded pedestrian detection task is handled in this paper by training with high quality samples and refining the bounding box regression. The visible IoU explicitly considers the visible region in selecting positive samples and effectively improves the training results. The sign predictor placed at the final stage detaches direction prediction from box regressor and enables the network to output more accurately localized boxes through extra training loss and refinement procedure. Our approach achieves state-of-the-art performance on CityPersons pedestrian detection benchmark. The proposed modules can be easily implemented and used for other object detection tasks and frameworks. 
\label{sec:concln}


\bibliographystyle{IEEEbib}
\bibliography{refs}

\end{document}